\pdfoutput=1

\documentclass[11pt]{article}

\usepackage{acl}

\usepackage{times}
\usepackage{amsthm}
\usepackage{latexsym}
\usepackage{graphicx}
\usepackage{float}
\usepackage{color}
\usepackage{caption}
\usepackage{subcaption}
\usepackage{bbm}
\usepackage{url}
\usepackage{amssymb}
\usepackage{amsmath}
\usepackage{multirow}
\usepackage{booktabs}
\usepackage{algorithm}
\usepackage{algpseudocode}
\usepackage{wrapfig}
\newtheorem{theorem}{Theorem}[section]

\newcommand{\our}{\textsc{CapEEN}}

\title{\textsc{CapEEN}: Image Captioning with Early Exits and Knowledge Distillation}

\author{Divya Jyoti Bajpai and Manjesh Kumar Hanawal\\
   Department of IEOR, IIT Bombay \\
  \texttt{\{divyajyoti.bajpai, mhanawal\}@iitb.ac.in}}

\begin{document}
\maketitle
\begin{abstract}
Deep neural networks (DNNs) have made significant progress in recognizing visual elements and generating descriptive text in image-captioning tasks. However, their improved performance comes from increased computational burden and inference latency. Early Exit (EE) strategies can be used to enhance their efficiency, but their adaptation presents challenges in image captioning as it requires varying levels of semantic information for accurate predictions. To overcome this, we introduce \our{} to improve the performance of EE strategies using knowledge distillation. Inference in \our{} is completed at intermediary layers if prediction confidence exceeds a predefined value learned from the training data. To account for real-world deployments, where target distributions could drift from that of training samples, we introduce a variant A-\our{} to adapt the thresholds on the fly using Multi-armed bandits framework. Experiments on the MS COCO and Flickr30k datasets show that \our{} gains speedup of $1.77\times$ while maintaining competitive performance compared to the final layer, and A-\our{} additionally offers robustness against distortions. The source code is available at \url{https://github.com/Div290/CapEEN}

\end{abstract}

\section{Introduction}\label{sec: intro}
Image captioning, a multifaceted challenge at the intersection of computer vision and natural language processing, has reaped the benefits of deep neural networks (DNNs), characterized by their increased scale and complexity.  This task entails not only the identification of visual elements within an image but also the intricate interpretation of their relationships. Notably, the encoder-decoder framework has made significant strides in sentence generation by anticipating the next word in a sequence \cite{anderson2018bottom, chen2021human, chen2015mind, fei2021memory, huang2019attention, vinyals2015show, xu2021towards, xu2015show, yao2018exploring, zhang2021rstnet, li2022blip, li2023blip}. This predictive modeling considers both the image's content and the preceding partial sentence, resulting in substantial progress in the field \cite{bai2018survey}. However, their large size restricts deployment in resource-constrained scenarios requiring fast inference. Early Exit (EE) strategies have emerged as a strategic solution to overcome this challenge.

%

\begin{figure}
    \centering
    \includegraphics[scale = 0.61]{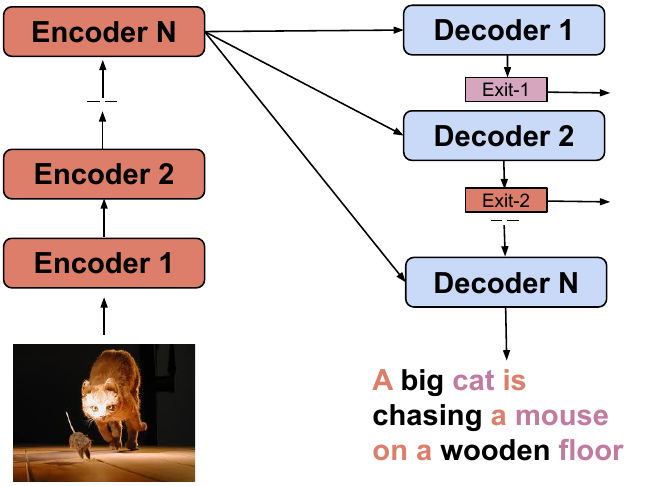}
    \caption{The encoder-decoder framework with attached exits. The figure states that low-level features could be extracted from early classifiers and inferred there, while high-level features are inferred at deeper classifiers. The color of the text in the caption is the same as the color of the classifier after that layer.}
    \label{fig:main figure}
\end{figure}

In EE strategies, classifiers are attached to the intermediary layers. Each sample can exit from one of them without requiring to pass through all layers (see Fig.~\ref{fig:main figure}). This brings down computational requirements and improves latency \cite{teerapittayanon2016branchynet, xin2020deebert}. However, this generic approach of attaching exits to the pre-trained backbone may not be suitable for image captioning \cite{fei2022deecap}-- in the layered hierarchy of representation in transformer-based models, the initial layers focus on extracting low-level features, while deeper layers delve into the complexities of semantic fusion relations \cite{cornia2020meshed, liu2021swin}. Consequently, even `easy samples' require a certain level of high-level information present at deeper layers.

Moreover, the decisions of early exit are based on the confidence at the intermediary layers being above a predefined threshold. The threshold used to compare the confidence levels significantly impacts the amount of latency and accuracy. These thresholds are learned during training and serve as a crucial reference point during inference.

\begin{figure}
    \centering
    \includegraphics[scale = 0.35]{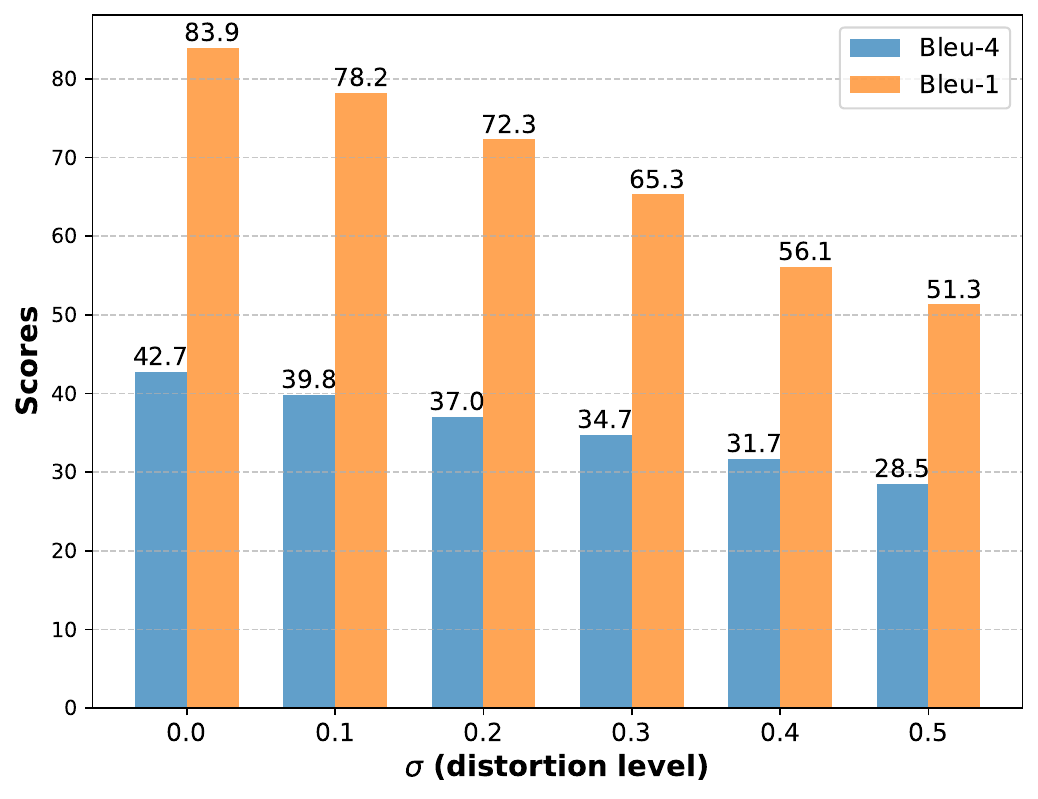}
    \caption{This figure shows the effect of distortion in the performance when the model was trained on undistorted images and tested on images with varying distortion levels ($\sigma$ models the distortion level).}
    \label{fig:noise_ki_effect}
    \vspace{-0.5cm}
\end{figure}

Post-deployment, it may be possible that the distribution of the target sample could drift away from that of training samples. This is often encountered in real-world scenarios, e.g., blurred images due to the camera being out of focus \cite{dodge2016understanding} during inference. Such drifts could affect the threshold choice and significantly lower the DNNs performance (see figure \ref{fig:noise_ki_effect}) prompting the question: How to adjust the threshold of deployed pre-trained models when the latent distribution of target samples differs from the training samples due to variation in distortion levels? Also, this adaptation has to be unsupervised, as the ground truth labels may not be available during inference.
This motivates a method that 1) gives initial layers high-level information during training and 2) adjusts early exit thresholds during inference for efficiency and robustness against distortions.

We introduce a new approach that extends the idea of knowledge distillation in early exits \cite{phuong2019distillation, zhu2021leebert} to image captioning tasks named Image \underline{Cap}tioning with \underline{E}arly \underline{E}xits and K\underline{N}owledge Distillation (\our{}) to improve the efficiency of EEs in image captioning tasks. 
By distilling the knowledge residing in deeper layers (teacher), \our{} empowers initial classifiers (students) to utilize the richness of deeper representations, which enhances both the performance and speed of the EE model.

To circumvent the issue of threshold choice under distribution change due to distortion in incoming samples during inference, we propose a novel online learning algorithm A-\our{} based on the Multi-Armed Bandits (MAB) framework \cite{auer2002finite} to learn the optimal threshold as per latent distributions of input samples. A-\our{} is activated during the inference phase and adapts to various levels of distortions in test samples with minimal computational requirements, making our method more suitable for real-world scenarios.



Our experiments on the MS COCO \cite{lin2014microsoft} and Flickr30k \cite{plummer2015flickr30k} dataset demonstrate that \our{} significantly increases speedup while maintaining competitive accuracy performance ($1.77$x speedup in runtime as compared to the final layer). A-\our{} further enhances efficiency by dynamically adjusting exit thresholds based on the latent distribution of the test dataset to make it robust to noise present in the datasets during inference. 

In summary, our contributions are as follows: 1) We present a novel self-distillation framework named \our{} tailored for early exiting in image captioning. 2) We introduce an online learning algorithm, A-\our{}, designed to dynamically choose optimal thresholds as per the latent data distribution utilizing the MAB framework. The algorithm uses the confidence scores to learn the optimal threshold. It makes the method robust to different levels of distortion in the test samples. 3) Comprehensive experiments conducted on the MS-COCO and Flickr30k dataset, reveal the superior performance of \our{} across all key metrics compared to previous methodologies. Additionally, we validate the effectiveness of A-\our{} in consistently determining optimal threshold values for various data distributions.

\section{Related works}
\textbf{Image Captioning:} Recent research has witnessed a surge in exploring efficient text description generation for input images \cite{bai2018survey}. Encoder-decoder frameworks have gained prominence for their exceptional performance in text generation tasks, leveraging contextual information \cite{anderson2018bottom, chen2021human, chen2015mind, fei2021memory, huang2019attention, vinyals2015show, xu2021towards, xu2015show, yao2018exploring, zhang2021rstnet, li2022blip, li2023blip}. With the advent of the attention mechanism \cite{vaswani2017attention}, the focus has shifted towards employing multiple layers of transformers as both encoders and decoders.\\
\textbf{Early-exits.} In recent years, the issue of inference latency has gained substantial attention \cite{matsubara2022split, guo2020non}. To address this issue, DNNs are implemented with internal classifiers in the intermediate layers. Notably, BranchyNet \cite{teerapittayanon2016branchynet} explores early classification at intermediate layers for images, while SPINN \cite{laskaridis2020spinn} uses a mobile cloud setup to split the DNN.  SEE \cite{wang2019see} performs the early exiting in a service outage scenario. Multiple early exiting frameworks have also been proposed such as \cite{huang2017multi, yang2020resolution, han2023dynamic} for improving early exits for image tasks dynamically choosing the depth of network for different regions for an image. Other works like \cite{phuong2019distillation} have utilized knowledge distillation in early exit framework but not for image captioning. Also, it performs joint training of teacher and student that deteriorates the optimality of the backbone.

Numerous early exiting frameworks have been devised for natural language processing tasks \cite{xin2020deebert, liu2021towards, liu2020fastbert, wang2019dynexit, li2021accelerating, zhou2020bert, zhu2021leebert, ji2023early, balagansky2022palbert, zhang2022pcee}, primarily based on the BERT backbone. DeeCap 
\cite{fei2022deecap} introduces early exiting to image captioning, employing an imitation network to replicate outputs from computationally intensive transformer layers within an encoder-decoder architecture. Similarly, MuE \cite{tang2023you} applies early exits to OFA \cite{wang2022ofa}, a unified vision language model designed for multi-modal applications.\\
\textbf{Multi-armed bandits in Early exits.} Several works utilize the MAB framework to adapt to different scenarios. Notable methods like LEE \cite{ju2021learning}, DEE \cite{ju2021dynamic}, and AdaEE \cite{pacheco2023adaee} aim to learn the optimal exit points in scenarios like mobile devices with restricted computational resources. Additionally, EPNet \cite{dai2020epnet} adopts an offline approach to learning when to exit based on considerations of computational overhead and accuracy. On the other hand, UEEUCB \cite{hanawal2022unsupervised} employs a Multi-Armed Bandit (MAB) framework to dynamically learn the optimal exit strategy in an online and unsupervised manner. UEEUCB relies on the assumption of strong dominance in neural networks, wherein accuracy increases with the layer number in the neural network.

The key differences are: 1) to the best of our knowledge, improving the performance of early exits using knowledge distillation has not been studied for image captioning. 2) Our online algorithm based on the MAB setup overcomes the challenge of choosing the optimal threshold by adapting to the underlying latent distributions of the noise present in the test dataset. 

\textbf{Note:} We are the first to apply knowledge distillation for image captioning. Previous methods have applied knowledge distillation for early exits in text and image classification which are much simpler tasks than image captioning and they simultaneously perform distillation in a single-stage training. This cannot be extended to image captioning as it requires high-quality knowledge transfer else it can lead to incorrect learning paths.

Given the complexity of image captioning, we perform a two-stage training that not only transfers high-quality knowledge but also maintains the optimality of the backbone. This approach provides us with a state-of-the-art backbone and serves as a testbed for A-\our{}.
\section{Methodology}
In this section, we discuss our method of fusing knowledge distillation with early exit layers by treating the final layer classifier as the teacher and the early exit classifiers as the students.
\subsection{Backbone}
We use the encoder-decoder framework for building our backbone network motivated by previous works \cite{liu2021cptr, li2023blip}. The encoder component comprises a pre-trained Swin-Transformer-base model \cite{liu2021swin}. What sets the Swin-Transformer backbone apart from other vision transformer models \cite{ranftl2021vision} is its Window and Shifted-Window Multi-head Self-Attention (SW-MSA) for the extraction of high-quality rich features. Swin's unique approach has consistently delivered state-of-the-art performance in various vision-related tasks \cite{wang2022end}. The encoder extracts rich features from the input image and enhances them by capturing their intra-relationships. The output of the Swin-Transformer encoder represents the image in a way that takes into account both local and global context. These features capture details, objects, and their spatial relationships within the image.
On the other hand, the decoder component uses the pre-trained GPT-2 \cite{lagler2013gpt2} model to generate captions in an autoregressive manner, effectively capturing the inter-relationships between words and image features. 

\subsection{Finetuning Backbone and Training Exits}
\our{} requires two main training steps: (i) The backbone fine-tuning and (ii) Training of the attached exits using knowledge distillation.
\subsubsection{\our{} backbone fine-tuning}
We start with a pre-trained encoder and decoder. The grid features of the image from the encoder output are passed to the decoder for cross-attention computation. The encoder-decoder backbone is then updated using cross-entropy loss calculated between the predicted token $y_i$ and the ground-truth token $y_i^{*}$. The loss function for fine-tuning is formulated as:
\begin{equation}
\nonumber
\mathcal{L}(I;\theta) = -\frac{1}{T}\sum_{t=1}^{T}\text{log}P_{N}(y_t^*|y_{1:t-1}^*, I; \theta),
\end{equation}
where $\theta$ denotes the collection of all the parameters, $I$ denote the input image, $T$ is the caption length, $y_{1:T}^{*}$ denotes ground-truth caption, $P_N$ denotes the probability score from the final layer, and $N$ denotes the number of layers in the decoder. We define vocabulary $\mathcal{V}$ as the set of tokens.
Once the fine-tuning is complete,  we freeze all the backbone parameters. This maintains the optimal quality of the backbone after exits are attached.

\begin{figure}
    \centering
    \includegraphics[scale=.61]{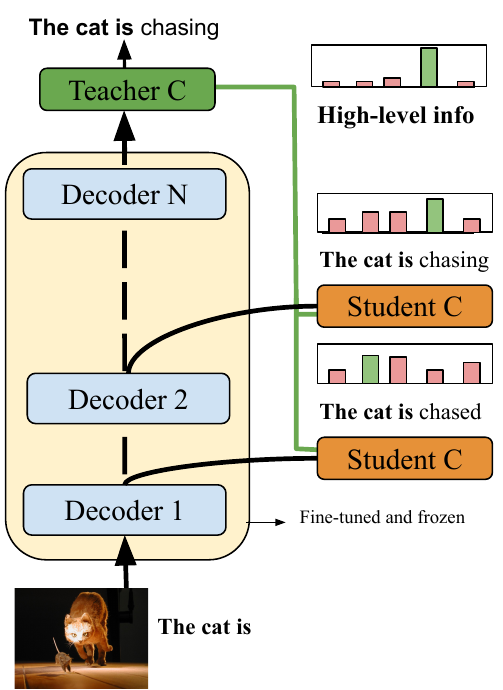}
    \caption{The overall training process for the decoder. Teacher C: Teacher classifier, Student C: Student Classifier, the bars show the probability distribution across different exits.}
    \label{fig:exits}
\end{figure}
\subsubsection{\our{} Exits Training}
After obtaining the fine-tuned backbone from the previous step, we attach task-specific exits at each decoder layer except the final layer. We use a student-teacher setup where the teacher is the final layer, and each intermediate classifier is treated as a student, as visualized in Fig. \ref{fig:exits}. The weights of the $i$th exit/student are trained using the loss:
\begin{multline}
\nonumber
\noindent
 \mathcal{L}_{i}(I; \theta, \theta_e) = \\-\frac{1}{T}\sum_{t=1}^{T}(\text{log}(P_i(y_t^*|y_{1:t-1}^{*}, I; \theta, \theta_e) \\
 + KL(p_t^i, p_t^n))
\end{multline}
where $\theta_e$ are the learnable weights for the exits, $y_t^{*}$ is the $t$th ground-truth token. $p_t^i$ is the probability vector on the vocabulary for $i$th student. Its $v$th component is given by $p_t^i(v) =  P_i({v}|y_{1:t-1}^{*}, I; \theta, \theta_e)$ and $p_t^n$ is the probability vector of the teacher model where $p_t^n(v) = P_N({v}|y_{1:t-1}^{*}, I; \theta)$. $KL$ is the KL-Divergence function defined as $KL(p_t^i, p_t^n) = \sum_{v\in\mathcal{V}}p_t^i(v)\log\frac{p_t^i(v)}{p_t^n(v)}$. The early classifiers are then jointly trained using the loss function $\sum_{i = 1}^{n-1}\mathcal{L}_{i}$. In this way, the learning is guided by hard and soft labels from the final layer. 

\subsection{\our{} inference}
We predict the caption in an autoregressive manner. This entails making token-by-token predictions for a given image, where the layer at which a token is predicted is determined by the prediction confidence 
$C_i = \max_{v\in \mathcal{V}} P_i(v | y_{1:t-1}, I; \theta, \theta_e)$. The input to the decoder 
is processed sequentially through the decoder layers until $C_i$ (the confidence value) is greater than a predefined threshold value $\alpha$. This threshold is set using the validation data based on the required accuracy-efficiency trade-off. The pseudo-code for inference is given in Algorithm \ref{alg: inference}. In the algorithm $<bos>$ and $<eos>$ denote the beginning of sentence and end of sentence token, respectively.  We rely on fixed confidence threshold values during inference 
that are tuned during training and applied uniformly across all exits to make decisions of early inference. We denote the prediction word by $c$ and the predicted caption by $\mathbf{C}$.
\begin{algorithm}
\caption{\our{} Inference}\label{alg: inference}
\begin{algorithmic}[1]
\State\textbf{Input:} Image $I$, Vocabulary $\mathcal{V}$, threshold $\alpha$.

\State $c = <bos>$, $\mathbf{C} = [<bos>]$
\While{$c \neq <eos>$}
\For{$i \gets 1 $ to $N$}
\State $C_i\gets \max_{v\in\mathcal{V}} P_i(v | I, \mathbf{C};\theta, \theta_e)$
\If{$C_i\geq\alpha$ and $i<N$}
\State $c = \arg\max_{v\in\mathcal{V}} P_i(v | I, \mathbf{C};\theta, \theta_e)$
\ElsIf{$i = N$}
\State $c = \arg\max_{v\in\mathcal{V}} P_N({v} | I, \mathbf{C}; \theta)$
\EndIf
\EndFor
\State $\mathbf{C}.append(c)$
\EndWhile
\State \textbf{Return:} $\mathbf{C}$
\end{algorithmic}
\end{algorithm}

\section{Learning of Thresholds}
Recall the discussion from the introduction that a threshold set using the validation set may not result in better performance when test data distribution drifts from that of the train data distributions, especially when the images in the test data are distorted. As the threshold used in early exit decisions significantly impacts both computational requirements and accuracy, setting it appropriately as per the test data distribution is crucial for optimal performance. We address this challenge by adjusting the threshold as per input data distribution in an online fashion using the MAB framework \cite{auer2002finite}.

In the MAB setup, a decision-maker repeatedly selects from a set of arms (or actions) while adapting to the unknown environment. Each arm corresponds to a specific choice or decision. The challenge lies in learning which arms yield the most favourable outcomes (highest reward) over time. This adaptation and learning process, central to MAB setups, aligns with the dynamic nature of online learning problems, where decisions are made sequentially based on incoming data. By leveraging the principles of exploration and exploitation, MAB frameworks facilitate learning the optimal action for the latent distribution. 

In this context, we define the action set as $k$ thresholds for each exit as $\mathcal{A} = \{\alpha_1, \alpha_2, \ldots, \alpha_k\}$. For exit $i$, we define the latency factor as the cost of processing the sample from the $1$st exit to the $i$th exit and denote it as $o_i$. Let $[N]$ denote the set $\{1, 2, \ldots, N\}$. A token will exit the backbone only if the confidence is above the chosen threshold. For a given threshold $\alpha$ where $\alpha \in \mathcal{A}$, suppose that $C_j<\alpha$ for $j \in [i-1]$ and $C_i\geq \alpha$ then it exits at $i$th exit, and the reward is defined as:
\begin{equation}
    r(\alpha) = (C_i - C_1)  - \mu o_i
\end{equation}
where $\mu$ is the scaling factor/conversion factor to bring the cost in terms of confidence. If the token does not gain sufficient confidence till the final layer, then $C_j<\alpha$ for all $j\in[N-1]$, and the token will be inferred at the final layer. Then the  reward is:
\begin{equation}
    r(\alpha) = (C_N - C_1) - \mu o_N
\end{equation}
The reward could be interpreted as follows: if a sample exits at layer $i$, then the gain is the confidence gain in the inference at layer $i$ compared to the inference at the first layer, and the cost incurred in processing the token from the first exit to the $i$th exit. The reward is the net gain, expressed as the difference between gain and cost. 
The objective is to maximize the expected reward as :

\begin{multline}
    \mathbb{E} [r(\alpha)] = \sum_{i = 1}^{N-1}\mathbb{E}[(C_i - C_1) - \mu o_i | \text{exit($i$)} ]\\.P[\text{exit($i$)}] + 
    \mathbb{E}[(C_N - C_1) - \mu o_N | \text{exit($N$)}]\\.P[\text{exit($N$)}] 
\end{multline}
where exit($i$) denotes the exit from $i$th layer. The objective is to find an action that maximizes the expected reward function. Note that the reward does not use any label information.  The optimal arm is defined as $\alpha^{*} = \arg\max_{\alpha\in\mathcal{A}}r(\alpha)$. If we consider a policy $\pi$ that selects threshold $\alpha_t\in \mathcal{A}$ in round $t$ based on past observations. The efficiency of the chosen policy can be expressed in terms of cumulative regret, defined as:
\begin{equation}
    R(\pi, T) = \sum_{t = 1}^{T}\mathbb{E}[r(\alpha^*)-r(\alpha_t)]
\end{equation}
\subsection{Algorithm}
\begin{algorithm}[H]
\caption{A-\our{}}\label{alg: gen_algorithm}
\begin{algorithmic}[1]
\State\textbf{Input:} $o_i \text{ }\forall i \in [N], \gamma\geq1$, $\mathcal{A}$\\
\textbf{Initialize:}  For an image obtain $|\mathcal{A}|$ tokens by setting different $\alpha \in \mathcal{A}$ and observe $r(\alpha)$. 
\State Set $Q({\alpha})\gets r(\alpha), N(\alpha)\gets 1, \forall \alpha$.
\State $t = |\mathcal{A}|+1$
\For{$j>1$}
\State For image $I_j$, set $c=\mathbf{C}_j=<bos>$

\While{$c\neq <eos>$}
\State $S_j = \{\mathbf{C}_j, I_j\}$
\State For next token, set threshold  
\State $\beta_t \gets \displaystyle \arg \max_{\alpha\in \mathcal{A}}\left(Q(\alpha)+\gamma\sqrt{\frac{\ln(t)}{N(\alpha)}}\right)$
\State $i = 1$ and $o_1 = 0$
\For{$i = 1 \textbf{ to } N$}
\State Pass $S_j$ till layer $i$ 
\State Apply threshold $\beta_t$ and observe $C_i$
\If{$C_i\geq \beta_t$ and $i < N$} 
\State  Infer at layer $i$ and exit
    \State $r_{t}(\alpha) \gets (C_i - C_1) -\mu o_i$ 
    \State $N_{t}(\alpha) \leftarrow N_{t-1}(\alpha)+1$
    \State $Q_{t}(\alpha) \leftarrow \frac{\sum_{j=1}^{t}r_{j}(\alpha_j)\mathbbm{1}_{\{\alpha_j=\alpha\}}}{N_{t}(\alpha)}$
    \State {\small $c = \arg\max_{v\in\mathcal{V}} P_i(v|S_j;\theta, \theta_e)$}
\State \textbf{break}

\ElsIf{$i = N$}
    \State Process till the last layer.
    \State $r_{t}(\alpha) \gets (C_N-C_1) - \mu o_N$ 
    \State $ N_{t}(\alpha) \leftarrow N_{t-1}(\alpha)+1$
    \State $Q_{t}(\alpha) \leftarrow \frac{\sum_{j=1}^{t}r_{j}(\alpha_j)\mathbbm{1}_{\{\alpha_j=\alpha\}}}{N_{t}(\alpha)}$
    \State $c = \arg\max_{v\in\mathcal{V}} P_N(v | S_j;\theta)$
\EndIf
\State $t\gets t+1$
\EndFor
\State $\mathbf{C}_j.append(c)$
\EndWhile
\State Return $\mathbf{C}_j$, $j\gets j+1$
\EndFor
\end{algorithmic}
\end{algorithm}
We introduce an algorithm called Adaptive-\our{} (A-\our{}) to adaptively choose the threshold values, and its pseudocode is outlined in Algorithm \ref{alg: gen_algorithm}. The input for this algorithm includes latency factors $o_i$ for each exit $i\in[N]$, the exploration parameter $\gamma$. 
The expectation above is with respect to the randomness induced by latent sample distribution in the selection of actions. A policy denoted as $\pi^{*}$ is characterized as sub-linear when the average cumulative regret diminishes, that is, $R(\pi^{*}, T)/T\rightarrow 0$. Our primary goal is to devise a learning algorithm that has a sub-linear regret guarantee.

The initialization of the algorithm involves ensuring that each action is played at least once. In the subsequent rounds, the algorithm selects the arm with the highest UCB index, denoted as $\beta_t$ (line 5). UCB indices comprise weighted averages of rewards $Q_t(\alpha)$ and incorporate confidence bonuses with $\gamma$ as the exploration parameter (weight). This confidence threshold is associated with each exit i.e., $\beta_t$ is the threshold chosen for all exits for the $t$th token. The token is then processed until the confidence is above $\beta_t$ (line 10) and if the token never gains sufficient confidence, it is inferred at the final layer (line 16). After the token exits from one of the layers, the average reward of the played arm is updated (line 14). The caption starts with a $<bos>$ token and the next predicted token $c$ is appended to the predicted caption set $\mathbf{C}$ until the $<eos>$ token is predicted. Note that there are two counters, one is for the image, once the $<eos>$ token is predicted the caption for the image is complete and the counter for the image is updated while the token counter is updated every time a token is passed through it and denotes the number of times rewards are updated. Hence, the algorithm learns faster as the thresholds (arms) are learnt on the number of tokens passed instead of the number of images.


Drawing from the analysis of UCB1 \cite{auer2002finite}, the regret of A-\our{} can be shown to be of the order $\mathcal{O}\left(\sum_{\alpha \in \mathcal{A}\backslash\alpha^{*}}\frac{log(n)}{\Delta{\alpha}}\right)$, where $\Delta_{\alpha} = r(\alpha^{*})-r(\alpha)$ denotes the sub-optimality gap. For completeness, we provide proof in the Appendix \ref{sec: regret bound}.

\section{Experiments}\label{sec: experiments}
\begin{table*}[]
\centering
\small
\begin{tabular}{cccccccc}
\hline
\textbf{Models/Metric }       & \textbf{BLEU-1}        & \textbf{BLEU-4}        & \textbf{METEOR}        & \textbf{CIDEr}          & \textbf{SPICE}         & \textbf{ROUGE-L}       & \textbf{Speedup }          \\ \hline
Final-Exit    & 82.9          & 42.3          & 32.2          & 147.1          & 26.7          & 61.3          & 1.00$\times$              \\ 
Decoder-9L       & 76.5          & 37.1          & 29.3          & 134.8          & 23.2          & 57.9          & 1.33$\times$          \\ \hline
DeeBERT       & 70.1          & 32.3          & 26.9          & 110.2          & 20.9          & 50.7          & 1.35$\times$          \\
ElasticBERT       & 71.4          & 32.8          & 27.6          & 114.6          & 21.4          & 51.6          & 1.37$\times$          \\
F-PABEE         & 72.7          & 33.9          & 27.9          & 115.6          & 21.9          & 52.3          & 1.30$\times$          \\
FastBERT       & 75.0          & 35.6          & 28.2          & 119.5          & 22.1          & 53.7          & 1.42$\times$          \\
LeeBERT       & 77.3          & 38.7          & 29.4          & 129.2          & 23.0          & 55.9          & 1.39$\times$          \\
DeeCap        & 77.5          & 39.2          & 29.9          & 132.8          & 23.2          & 56.9          & 1.60$\times$          \\
MuE           & 79.3          & 40.5          & 30.9          & 139.4          & 24.9          & 59.7          & 1.66$\times$          \\ 
\textbf{Ours} & \textbf{80.7} & \textbf{41.3} & \textbf{31.6} & \textbf{140.3} & \textbf{25.5} & \textbf{60.1} & \textbf{1.77$\times$} \\ \hline
\end{tabular}
\caption{Main Results on COCO dataset: CapEEN outperforms other baselines across different metrics.}
\label{tab: results}
\end{table*}

\textbf{Dataset and Metric:} We evaluate the performance of our method using the MS-COCO and Flickr30k datasets for image captioning. Our primary objective is to produce coherent image captions. To maintain consistency with prior studies, we preprocess all captions by converting them to lowercase. Additionally, we filter out words that occur fewer than 5 times in the dataset, ensuring robustness in our evaluation.
We report key metrics, including BLEU-4 \cite{papineni2002bleu}, METEOR \cite{banerjee2005meteor}, CIDEr \cite{vedantam2015cider}, ROUGE \cite{lin2004rouge} and SPICE \cite{anderson2016spice} scores. To be consistent with previous methods, we report the speedup ratio as the measure of reduction in computational requirements. This metric can easily be converted to the expected time reduction rate.

\begin{equation}
    \frac{\sum_{l=1}^{N}w_l^I\times N}{\sum_{l=1}^{N} w_l^I\times l}
\end{equation}
where $N$ is the number of decoder layers and $l$ is the number of layers after which the token exits the backbone during inference. $w_l^I$ is the number of words that exit at the $l$th decoder layer for an image $I$. This metric provides insights into our decoding process's resource utilization and efficiency, a critical aspect discussed in our work.

\noindent
\textbf{Training:}
The encoder-decoder backbone is initially fine-tuned for 10 epochs with a starting learning rate of 1e-5, which decays by 0.5 every 3 epochs. Subsequently, self-critical training is employed for 5 epochs with an initial learning rate of 7e-6, also decaying by 0.5 every 3 epochs. The backbone weights are frozen post-fine-tuning, and exits are added to the decoder, whose weights are further trained for 5 epochs. The Adam optimizer and a batch size of 8 are chosen, with a threshold of 0.6 chosen based on the accuracy-efficiency trade-off on the validation split. Inference is conducted on the Karpathy test split of the MS-COCO dataset with a batch size of 1, using NVIDIA RTX 2070 GPUs. Results are presented in Table \ref{tab: results}.  More runtime details are in Appendix \ref{sec: computational}

\noindent\textbf{Adaptive threshold learning (A-\our{}):} We experiment with two types of noise, Gaussian noise and Gaussian blur. We mimic real-world scenarios by adding different levels of noise to the images of the test set and then proceed to learn the thresholds. In this, we adapt the thresholds based on the level of distortion present in an image using Algorithm \ref{alg: gen_algorithm}. For this step, we choose the action set as $\mathcal{A} = \{0.1, 0.2, \ldots, 1.0\}$. We add noise to the dataset's test split. Then we perform learning of threshold values, using A-\our{} for all the exits. Note that algorithm \ref{alg: gen_algorithm} has small computational complexity and does not add upon latency in the inference process. It maximizes the rewards over a finite set which has negligible complexity. 

We set $\mu = 1/N$. The latency cost $o_i$ could be understood as the cost of processing the samples from $1$st exit to $i$th exit. Hence we set the latency cost as $o_i = \lambda i$ where $\lambda$ is per layer computational cost. Since the value of threshold $\alpha$ is adaptively chosen in this case, $\lambda$ models the trade-off between accuracy and efficiency. A higher value of $\lambda$ will provide higher accuracy while a lower value will give high efficiency. We use the value of $\lambda = 1$. A detailed ablation study of this hyperparameter can be found in the Appendix \ref{sec: lambda}.

\noindent
\textbf{Impact of change in Distribution:} We evaluate the impact of data distribution shifts on our model's performance (BLEU-1, BLEU-4) using image distortion (Figure~\ref{fig:noise_ki_effect}). We train on undistorted MS-COCO images and introduce Gaussian noise/blur to the test set, simulating real-world imperfections. Noise levels are varied for Gaussian noise ($\sigma \in \{0.1, 0.2, 0.3, 0.4, 0.5\}$) and Gaussian blur ($\sigma \in \{0.5, 1.0, 1.5, 2.0\}$) to assess robustness to context-driven data variations.

\noindent
\textbf{Baselines:}  We establish baseline models for performance evaluation. To assess improvements in speedup as compared to the final decoding layer, we consider this setup as a baseline for our approach (final exit). In this case, all the samples are inferred only at the final layer. We also directly reduce the layer number to $9$ in Decoder-9L and use only $9$th layer to make an inference. This baseline serves as a lower bound for performance metrics since it does not employ any technique.
 Since DeeBERT \cite{xin2020deebert}, ElasticBERT \cite{liu2021towards}, F-PABEE \cite{zhou2020bert}, FastBERT \cite{liu2020fastbert} and LeeBERT \cite{zhu2021leebert} were originally conducted on BERT, we implemented their methods in the decoder part of GPT-2 without changing any hyperparameters.
We compare our model with DeeCap \cite{fei2022deecap} and MuE \cite{tang2023you} utilizing the setup from their framework. DeeCap implements an imitation network where as a sample exits the backbone, it passes through multiple MLP layers to regain lost information due to early exiting. MuE is another early exiting model for image captioning that attaches exits in the OFA backbone which is a multimodal unified vision language model. It has the similarity score of consecutive layers as the confidence metric. More details on baselines can be found in Appendix \ref{sec: baselines}. Note that except for DeeCAP, we have applied the methodology of other baselines to our setup.

We evaluate the performance of \our{} on the Karpathy test split of the MS-COCO dataset, as done by all the baselines for fair evaluation, and the results are in Table \ref{tab: results}. The results of the Flickr30k dataset are in Appendix \ref{sec: flickr} and table \ref{tab: res_flickr}.


\begin{table}[]
\small
\centering
\begin{tabular}{ccccc}
\hline
\textbf{Model} & \textbf{BLEU-4} & \textbf{METEOR} & \textbf{CIDEr} & \textbf{Speed} \\ \hline
\multicolumn{5}{c}{\textbf{Blur intensity = 0.5}}                                              \\ \hline
Final          & 39.2            & 31.6            & 148.7          & 1.00$\times$              \\
DeeCAP         & 34.7            & 28.4            & 133.6          & 1.33$\times$           \\
MuE            & 35.5            & 28.9            & 137.2          & 1.47$\times$           \\
Our            & 36.8            & 29.5            & 139.9          & 1.55$\times$           \\
A-our          & \textbf{37.3}   & \textbf{29.8}   & \textbf{141.8} & \textbf{1.61$\times$}  \\ \hline
\multicolumn{5}{c}{\textbf{Blur intensity = 1.0}}                                              \\ \hline
Final          & 37.9            & 30.7            & 139.5          & 1.00$\times$              \\
DeeCAP         & 33.2            & 27.5            & 126.3          & 1.31$\times$           \\
MuE            & 34.5            & 28.3            & 129.4          & 1.48$\times$           \\
Our            & 36.1            & 28.6            & 130.9          & 1.53$\times$           \\
A-our          & \textbf{37.0}     & \textbf{29.2}   & \textbf{132.7} & \textbf{1.59$\times$}  \\ \hline
\multicolumn{5}{c}{\textbf{Blur intensity = 1.5}}                                              \\ \hline
Final          & 34.1            & 28.6            & 127.4          & 1.00$\times$              \\
DeeCAP         & 28.4            & 26.8            & 113.5          & 1.28$\times$           \\
MuE            & 29.7            & 27.4            & 117.2          & 1.46$\times$           \\
Our            & 30.9            & 27.9            & 120.6          & 1.50$\times$            \\
A-Our          & \textbf{31.7}   & \textbf{28.5}   & \textbf{123.0} & \textbf{1.63$\times$}  \\ \hline
\multicolumn{5}{c}{\textbf{Blur intensity = 2.0}}                                              \\ \hline
Final          & 28.7            & 22.6            & 100.4          & 1.00$\times$              \\
DeeCAP         & 21.2            & 18.4            & 91.7           & 1.19$\times$           \\
MuE            & 23.9            & 20.3            & 94.5           & 1.35$\times$           \\
Our            & 25.4            & 21.0            & 98.2           & 1.39$\times$           \\
\textbf{A-Our} & \textbf{26.5}   & \textbf{21.9}   & \textbf{101.2} & \textbf{1.49$\times$}  \\ \hline
\end{tabular}
\caption{Results of A-\our{} when the test sample contains images with different levels of blur.}
\label{tab:blur_res}
\end{table}

\vspace{-0.1cm}
\subsection{Results}
In this section, We discuss the main results (median of $5$ independent runs) of our work. Details of the stability of our method are in table \ref{tab: stability}.

\textbf{\our{}.}
In the context of pristine (undistorted) images, Table \ref{tab: results} presents performance results of early exit models, showcasing our approach's superiority over various baselines. Our method outperforms all previous baselines, including DeeCap and MuE. Unlike DeeCap, our approach does not rely on an imitation network, avoiding noise accumulation as samples exit the main backbone early. Additionally, our method consistently outperforms MuE by leveraging deep representations crucial for semantic correctness. Baselines like DeeBERT, ElasticBERT, and F-PABEE exhibit significant performance drops due to limited access to deep representations. While FastBERT and LeeBERT performs better than these baselines by accessing deep representations, they lacks the appropriate ground-truth information during weight learning for attached exits. 

\our{} enriches early classifiers with higher-level semantic information distilled from the final layer, leading to minimal performance drops and the highest speedup ratio across all metrics.

\begin{figure*}
    \centering
    \begin{subfigure}{0.3\textwidth}
        \includegraphics[width=\textwidth]{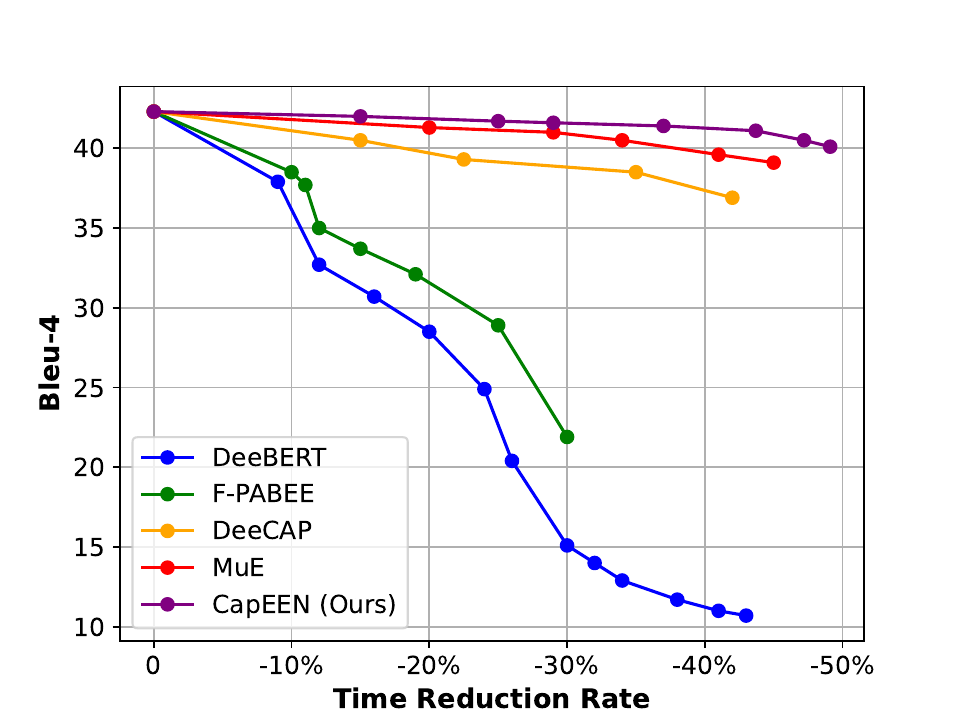}
        \label{fig:bleu}
    \end{subfigure}
    \begin{subfigure}{0.3\textwidth}
        \includegraphics[width=\textwidth]{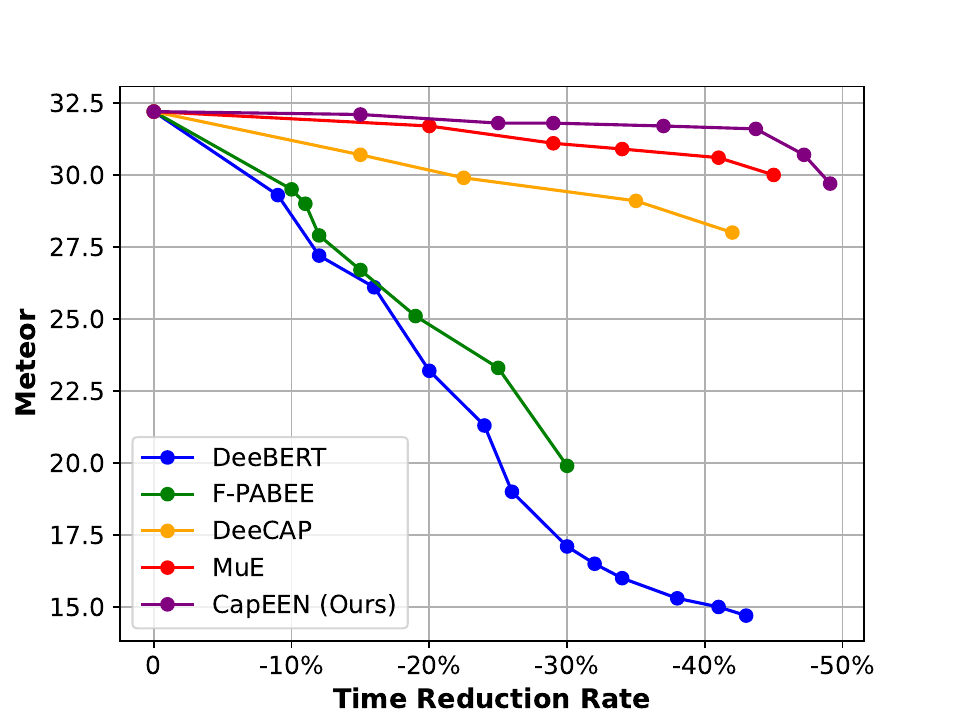}
        \label{fig:meteor}
    \end{subfigure}
    \begin{subfigure}{0.3\textwidth}
        \includegraphics[width=\textwidth]{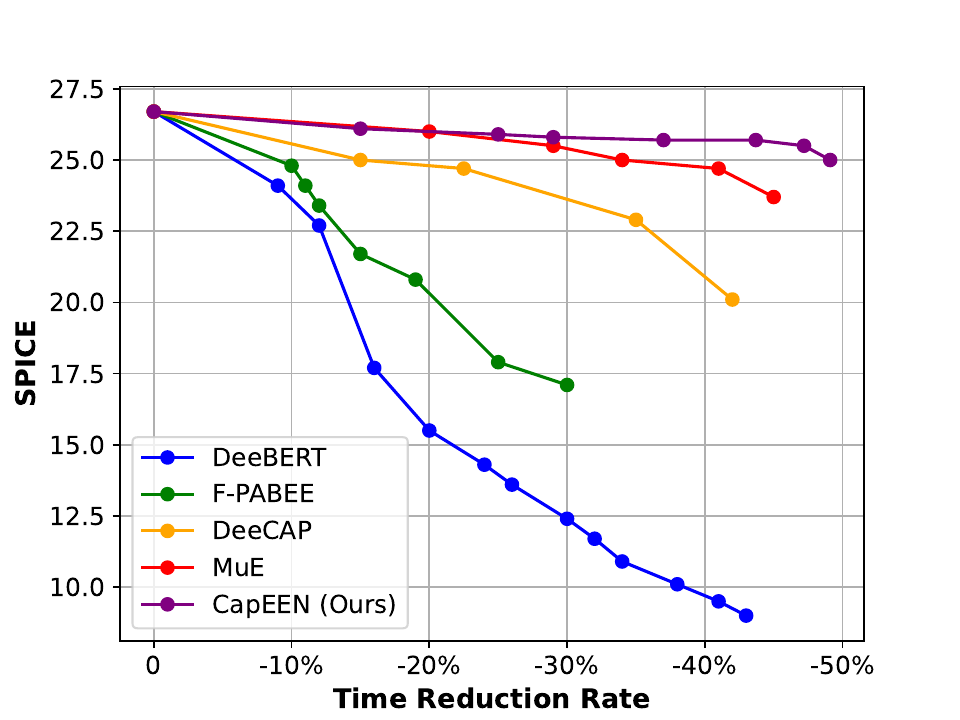}
        \label{fig:spice}
    \end{subfigure}
    \caption{Change in the performance of different metrics with changing time reduction rates. These reductions are observed by changing the threshold parameter $\alpha$.}
    \label{fig:global_fig}
\end{figure*}

\textbf{Adaptive learning of the thresholds (A-\our{}).}
In Table \ref{tab:blur_res} and \ref{tab: distortion res}, we highlight the impact of adapting threshold values based on changes in data distribution, comparing them with thresholds learned during training and fixed during inference. Our findings demonstrate that fixed thresholds significantly affect inference time and performance, highlighting the need for dynamic threshold learning to accommodate contextual information and inherent image noise.


For the CIDEr metric, A-\our{} observes minimal occasional gains from the final decoder layer, attributed to overthinking \cite{zhu2021leebert} during inference similar to overfitting during training. A-\our{} consistently outperforms \our{} as well as other baselines when the dataset distribution changes due to distortion in images. The gain in performance and speed is observed as A-\our{} finds the optimal threshold that optimally models the accuracy-efficiency trade-off.

We perform an ablation study and a case study in Appendix \ref{sec: more experiments} to further prove the effectiveness of our method.

\subsection{Analysis of threshold $\alpha$}\label{sec: threshold}
We present in the figure \ref{fig:global_fig}, the trade-off between accuracy and efficiency. To obtain higher accuracy, we increase the value of $\alpha$ and the time reduction decreases with higher accuracy. On the other hand, decreasing the threshold $\alpha$ will in turn increase the time reduction but with compromising performance. \our{} performed better than other baselines due to the available knowledge from deeper exits. Observe that in figure \ref{fig:global_fig}, there is a very minimal drop in performance (mostly constant). This ability comes from the extra information available at the intermediate exits due to knowledge distillation. However, the performance begins to drop when we try to reduce time by more than $50\%$ since then the sample exits from the very initial layers which in turn affects the performance.

\section{Conclusion}
\label{sec:Conclusion}
We introduced a new encoder-decoder backbone for image captioning with early exits and knowledge distillation named CapEEN. Using the student-teacher model, we trained early exit classifiers using knowledge distillation to capture high-level semantic representations available at the deeper layers. We demonstrated that CapEEN offers a significant increase in speedup while maintaining a competitive performance guarantee. Further, we introduced A-CapEEN, where the threshold used for early exit decisions can be adaptively learned for distributions that differ from the training data due to changes in the distortion levels.

\section{Limitations}

In our work, we have applied a uniform threshold across all exits during inference, simplifying the implementation process. However, extending this to individual thresholds for each exit could offer additional flexibility, albeit with increased complexity. Additionally, while early exit models effectively reduce latency during inference, they do incur higher computational costs during training. This is due to the need for exit classifiers to learn additional weights after each layer, resulting in increased complexity. Nonetheless, post-training, these models significantly enhance inference speed, which becomes the primary consideration post-deployment.

\section*{Acknowledgements}
Divya Jyoti Bajpai is supported by the Prime Minister’s Research Fellowship (PMRF), Govt. of India.  Manjesh K. Hanawal thanks funding support from SERB, Govt. of India, through the Core Research Grant (CRG/2022/008807) and MATRICS grant (MTR/2021/000645), and DST-Inria Targeted Programme. 
\bibliography{custom}

\appendix

\section{Appendix}
\label{sec:appendix}

\subsection{Regret Bound}\label{sec: regret bound}
\label{sec:proof}
\begin{theorem}
    For any $\gamma \geq 1$, the regret of A-\our{} with $K$ arms in the action set after $T$ rounds is given as:
    \begin{multline}
        R(A-\our{}, T)\leq 4\gamma \sum_{\alpha\neq\alpha^{*}}\frac{log(T)}{\Delta_{\alpha}}\\+\left(\frac{\pi^2}{3}+1\right)\sum_{\alpha\neq\alpha^{*}}\Delta_\alpha
    \end{multline}

where $\Delta_{\alpha} = r(\alpha^{*})-r(\alpha)$. 
\end{theorem}

\noindent \textbf{Proof.} The proof follows similar lines as given in the classical UCB \cite{auer2002finite}. the instantaneous regret in round $t$ is given as :
$$R_t = r(\alpha_t) - r(\alpha^{*})$$
The value of $r(\alpha)$ defined in our problem is a bounded quantity. More specifically $r(\alpha)\in [-1-\lambda N, 1]$ where $\lambda$ is the processing cost of the layers and $N$ is the number of layers in the decoder. 

The bound given in algorithm \ref{sec:proof} can be further improved by adding multiple updates for a single sample. To get an idea of the number of arms that will get updated, we provide the following proposition:

\section{More experimental details}\label{sec: more experiments}
\subsection{Ablation study}
In Figure \ref{fig:Ablation}, we conduct an ablation study by learning the early exit weights using different loss combinations. First, we learn the early exit only using the cross-entropy loss where only ground-truth labels are used to train the weights of early exits. Then we consider only knowledge distillation loss and only soft labels guide the early exits. 
We observe that if we do not use the combination of both soft as well as hard labels then early classifiers get the highest hit in terms of performance as they need rich information from deeper layers. As we move deeper into the backbone all three combinations converge to similar CIDEr scores as deeper layers already have access to rich features. 

These observations suggest the significance of the various components within the loss function, ultimately guiding us towards a more comprehensive understanding of the model's behaviour.

\subsection{Case Study}
\begin{figure*}
    \centering
    \begin{subfigure}{0.3\textwidth}
        \includegraphics[width=\textwidth]{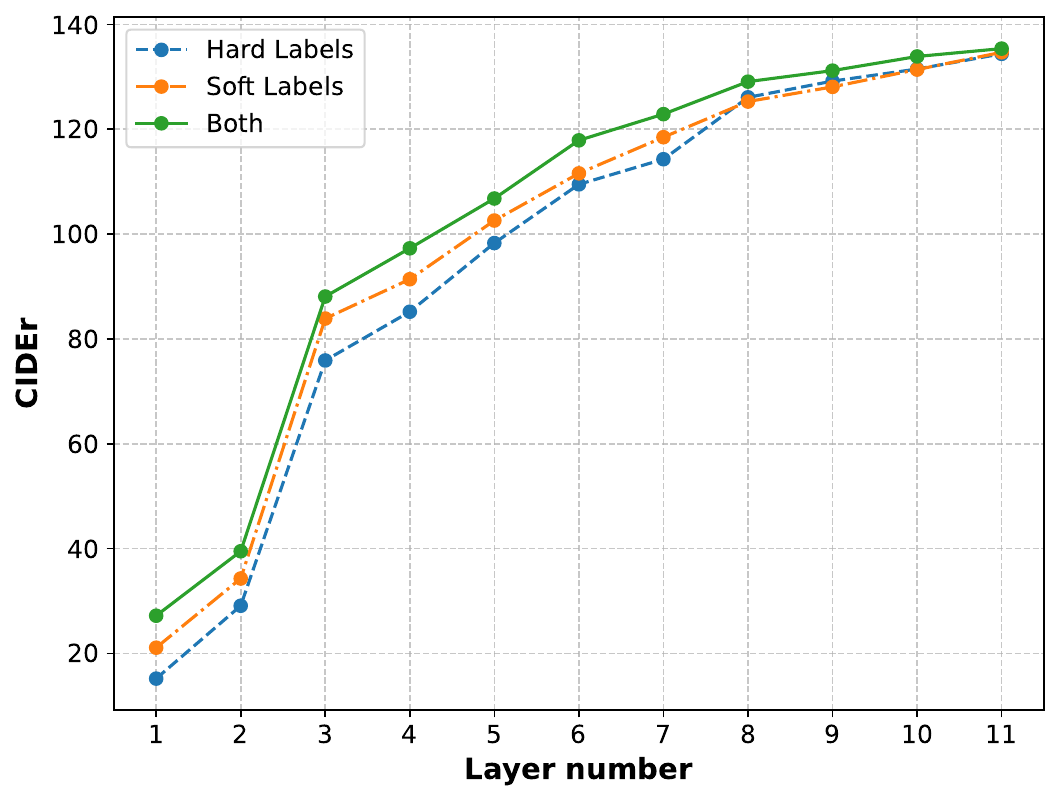}
        \caption{Effect of Loss Combinations on CIDEr Score across the layers.}
        \label{fig:Ablation}
    \end{subfigure}
    \begin{subfigure}{0.3\textwidth}
        \includegraphics[width=\textwidth]{ACL2024/figures/Noise_effect.pdf}
        \caption{Effect of different distortion levels on Bleu-1 and Bleu-4 metrics on final layer.}
        \label{fig:distortion}
    \end{subfigure}
    \begin{subfigure}{0.34\textwidth}
        \includegraphics[width=\textwidth]{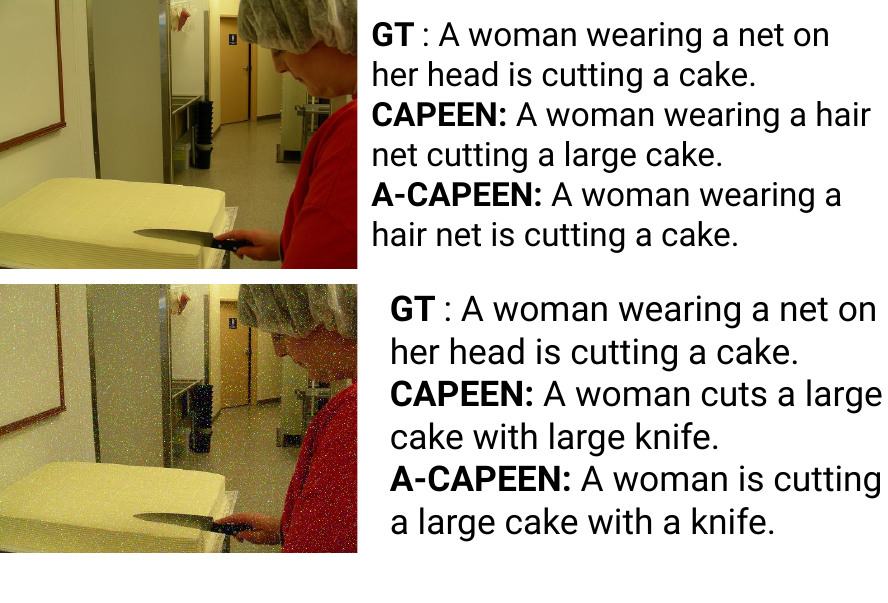}
        \caption{Predicted captions by \our{} and A-\our{} Ground-truth(GT) caption.}
        \label{fig:Caption}
    \end{subfigure}
    \caption{(a) Hard Labels (Ground-truth), Soft Labels (Teacher Model Predictions), and Both (Combined Labels). (b) Image is added with noise and then the captions are predicted, we report the Bleu-1 and Bleu-4 scores. (c) The first image is pristine (undistorted) while the second image is distorted with $\sigma = 0.2$ and the captions are predicted using \our{}and A-\our{}.}
    \label{fig:global}
\end{figure*}

We also provide some examples of how the different proposed models annotate an image. We compare the captions generated by the two proposed methods with the captions provided by humans (ground-truth). As we can observe from figure \ref{fig:Caption}, \our{} performs well in explaining the content of the image. As compared to the ground-truth, \our{} provides more details as it also recognizes that the cake is large. A-\our{} gets a caption which is closer to the \our{} when there is no noise in the image. However, as we add noise to the image, then the results of A-\our{} are better than \our{}. Note that A-\our{} is tested on this example after it has seen a sufficient amount (around 100 images) of data with similar noise so that it adopts the distribution of the data. Also, observe that there is a slight variation in the captions and some information is lost when there is added noise in the dataset. Still, the prediction given by A-\our{} is better as it has added 'is cutting' in the sentence while \our{} just outputs 'cuts' and \our{} observes that it is a large knife which is redundant information.

. 

\begin{figure}
    \includegraphics[scale = 0.35]{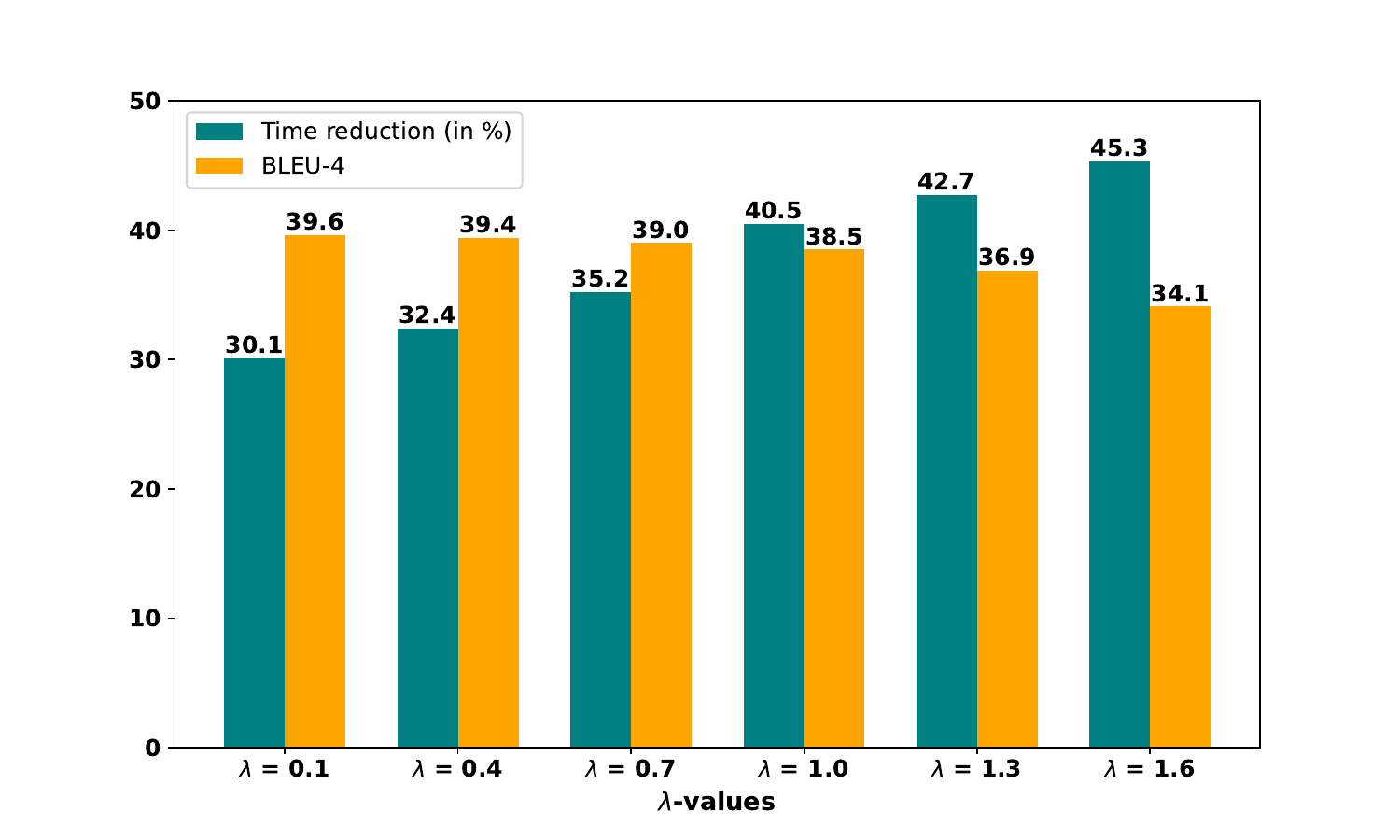}
    \caption{The change in time reduction rate as well as the BLEU-4 scores when the values of the $\lambda$ are varied.}
    \label{fig:analysis_lambda}
\end{figure}

\begin{table*}[]
\centering
\begin{tabular}{cccccc}
\hline
Model      & BLEU-4      & CIDEr      & SPICE       & ROUGE-L    & Speedup      \\ \hline
Final-exit & 42.7        & 154.1      & 26.7        & 63.3       & 1.00x        \\ \hline
CapEEN     & 41.3$\pm$0.015 & 140.3$\pm$0.3 & 25.5$\pm$0.009 & 62.3$\pm$0.07 & 1.77x$\pm$0.058 \\
A-CapEEN   & 42.1$\pm$0.08  & 144.2$\pm$1.1 & 26.1$\pm$0.08  & 63.0$\pm$0.5  & 1.76x$\pm$0.17  \\ \hline
\end{tabular}
\caption{This table shows the stability of \our{} and A-\our{} where $5$ independent runs are made.}
\label{tab: stability}
\end{table*}

\subsection{Baselines}\label{sec: baselines}
\noindent\textbf{DeeBERT}: do a separate training of the backbone and the attached exits in a setup similar to ours. It trains the model by only using the ground-truth labels and is originally trained for text classification tasks. We implement it in the decoder part of our backbone.

\noindent\textbf{ElasticBERT: } is similar to DeeBERT but performs a joint training of the backbone i.e. all the exits with the final exit are trained simultaneously. We implement it in the decoder part and provide the results.

\noindent\textbf{F-PABEE: } has different exiting criteria from DeeBERT and ElasticBERT, it exits the samples based on the consistency in the prediction from the intermediate classifiers being above a threshold. Its main purpose was to reduce the time as well as overthinking problems in DNNs.

\begin{table}[]
\small
\begin{tabular}{ccccc}
\hline
\textbf{Model}      & \textbf{BLEU-4} & \textbf{METEOR} & \textbf{CIDEr} & \textbf{Speed} \\ \hline
\multicolumn{5}{c}{\textbf{$\sigma$ = 0.5}}                                                 \\ \hline
\textbf{Final-exit} & 28.5            & 24.0              & 99.5           & 1.00x             \\
\textbf{Ours}       & 26.1            & 23.9            & 91.1           & 1.43x          \\
\textbf{A-Ours}     & \textbf{26.7}   & \textbf{24.1}   & \textbf{91.9}  & \textbf{1.51x} \\ \hline
\multicolumn{5}{c}{\textbf{$\sigma$ = 0.4}}                                                 \\ \hline
\textbf{Final-exit} & 31.7            & 26.5            & 113.9          & 1.00x             \\
\textbf{Ours}       & 28.9            & 26.4            & 105.2          & 1.53x         \\
\textbf{A-Ours}     & \textbf{29.2}   & \textbf{26.4}   & \textbf{107.9} & \textbf{1.59x} \\ \hline
\multicolumn{5}{c}{\textbf{$\sigma$ = 0.3}}                                                 \\ \hline
\textbf{Final-exit} & 34.7            & 28.9            & 122.1          & 1.00x             \\
\textbf{Ours}       & 32.7            & 28.3            & 115.8          & 1.50x          \\
\textbf{A-Ours}     & \textbf{33.5}   & \textbf{28.4}   & \textbf{121.7} & \textbf{1.58x} \\ \hline
\multicolumn{5}{c}{\textbf{$\sigma$ = 0.2}}                                                 \\ \hline
\textbf{Final-exit} & 37.1              & 30.7            & 134.8          & 1.00x             \\
\textbf{Ours}       & 35.7            & 30.1            & 127.3          & 1.51x         \\
\textbf{A-Ours}     & \textbf{35.8}     & \textbf{30.1}   & \textbf{129.4} & \textbf{1.58x} \\ \hline
\multicolumn{5}{c}{\textbf{$\sigma$ = 0.1}}                                                 \\ \hline
\textbf{Final-exit} & 39.8            & 31.9            & 142.8          & 1.00x             \\
\textbf{Ours}       & 38.5            & 31.1            & 132.3          & 1.62x         \\
\textbf{A-Ours}     & \textbf{38.8}   & \textbf{31.3}   & \textbf{135.1} & \textbf{1.68x} \\ \hline
\multicolumn{5}{c}{\textbf{$\sigma$ = 0.0}}                                                 \\ \hline
\textbf{Final-exit} & 42.3            & 32.2            & 147.1          & 1.00x             \\
\textbf{Ours}       & 41.3            & 31.6            & 140.3          & \textbf{1.77x} \\
\textbf{A-Ours}     & \textbf{41.8}   & \textbf{32.0}     & \textbf{142.6} & {1.76x}          \\ \hline
\end{tabular}
\caption{Comparison of \our{} and A-\our{} with different distortion levels.}
\label{tab: distortion res}
\end{table}

\noindent\textbf{FastBERT:} has also been originally implemented on the BERT backbone. For the first few epochs of training, it performs a joint training of the backbone and the exits and then more epochs on distilling the knowledge from deeper exits but without using the ground-truth. We implement it in the GPT-2 i.e. decoder part of our method. 

\noindent\textbf{LeeBERT} use prediction stability to decide early exits, LeeBERT also distil the knowledge from deeper layers. It perfoms cross level optimization to make the exits and the backbone learn better.

\noindent \textbf{DeeCAP: } is specifically made for early exits in image captioning. It learns an imitation network in the form of MLP(Multi-layer Perceptron) layers for each sample as it exits the backbone. Due to a separate MLP layer for each exit, it increase the size of the model to a greater extent. The expected time reduction rate is also affected as each time the sample has to check whether to exit a layer or not from the original backbone it has to exit from the main backbone and pass through the imitation network until the confidence is above a given threshold.

\noindent \textbf{MuE:} is an early exit mechanism in DNNs designed for making inferences at the intermediate classifier for the multi-modal tasks. The base model it has used is the OFA model. It sets the similarity in the hidden representations as the exiting criteria. This exiting criterion gives an advantage that it could also be applied to the encoder but has a restriction that the layers of the encoders are identical which might not be always the case \cite{liu2021swin, he2016deep}.

\subsection{Analysis of the cost $\lambda$}\label{sec: lambda}
In figure \ref{fig:analysis_lambda}, we perform analysis on the cost parameter $\lambda$. We introduced the cost parameter $\lambda$ as a user-defined parameter. It can also be understood as a trade-off factor between accuracy and efficiency. Since a higher value of $\lambda$ will increase the impact of processing cost and in turn will force samples to make an early exit by lowering the threshold values. This will increase the efficiency of the model but in turn, will decrease the accuracy of predictions. While a smaller value of $\lambda$ will motivate the samples to gain higher confidence by processing them to deeper layers. The latter method will increase the accuracy while compromising the efficiency of the model.
\begin{table*}[]
\centering
\begin{tabular}{cccccccc}
\hline
\textbf{Model/Metric} & \textbf{BLEU-1} & \textbf{BLEU-4} & \textbf{METEOR} & \textbf{CIDEr} & \textbf{SPICE} & \textbf{ROUGE-L} & \textbf{Speed}  \\ \hline
\textbf{Final-Exit}   & 76.9            & 33.6            & 25.5            & 72.7           & 18.1           & 55.3             & 1.00$\times$              \\ \hline
\textbf{DeeBERT}      & 65.2            & 24.5            & 20.8            & 47.5           & 12.6           & 45.1             & 1.24$\times$          \\
\textbf{ElasticBERT}  & 66.4            & 25.3            & 21.1            & 49.3           & 13.2           & 45.9             & 1.27$\times$         \\
\textbf{F-PABEE}      & 68.6            & 26.5            & 21.9            & 51.9           & 13.8           & 46.3             & 1.29$\times$          \\
\textbf{FastBERT}     & 70.1            & 27.9            & 22.8            & 55.3           & 14.7           & 48.5             & 1.33$\times$          \\
\textbf{DeeCap}       & 72.8            & 30.1            & 23.5            & 64.2           & 16.1           & 51.7             & 1.41$\times$          \\
\textbf{MuE}          & 74.2            & 31.9            & 24.4            & 66.7           & 16.9           & 53.4             & 1.58$\times$          \\ \hline
\textbf{Ours}         & \textbf{75.1}   & \textbf{32.8}   & \textbf{25.0}   & \textbf{67.2}  & \textbf{17.4}  & \textbf{53.9}    & \textbf{1.65$\times$} \\ \hline
\end{tabular}
\caption{Results on Flickr30k dataset}
\label{tab: res_flickr}
\end{table*}
\subsection{Computational requirements}\label{sec: computational}

We perform experiments on three NVIDIA RTX 2070 GPUs and the training time takes $\sim 10$ hours for the backbone fine-tuning and an additional $1$ hour for exits training. The model has $205$ million parameters and not all are updated as the backbone is pre-trained and not all parameters need to be updated as we are only fine-tuning it. The inference time is $<1$ hour on the Karpathy test split. When we also employ A-CapEEN i.e. we adapt the threshold values then the time required is the same as running CapEEN individually since the MAB framework does not add minimal computational complexity.

\section{Results on Flickr30k}\label{sec: flickr}
In table \ref{tab: res_flickr}, we provide the results on Flickr30k when the experiments are performed in a similar setup as given in the experimental section \ref{sec: experiments}. Our method consistently outperforms the other baselines in terms of all the metrics. Again this gain in performance with decreasing sufficient time is possible due to the higher level of information available at intermediate exit points. This encourages the exit of more samples early and with high confidence. Hence we prove across different datasets that \our{} outperforms previous baselines.

\end{document}